\documentclass[lettersize,journal]{IEEEtran}
\usepackage{amsmath,amsfonts}
\usepackage{enumitem}
\usepackage{algorithmic}
\usepackage{algorithm}
\usepackage{array}
\usepackage[caption=false,font=normalsize,labelfont=sf,textfont=sf]{subfig}
\usepackage{textcomp}
\usepackage{stfloats}
\usepackage{url}
\usepackage{verbatim}
\usepackage{graphicx}
\usepackage{multirow}
\usepackage{color}
\usepackage{todonotes}
\newtheorem{remark}{Remark}
\usetikzlibrary{positioning}
\usepackage{hyperref}
\usepackage[square, numbers, sort&compress]{natbib}
\begin{document}


\title{Traversability-Aware Legged Navigation by Learning from Real-World Visual Data}

\author{Hongbo Zhang\textsuperscript{1}, Zhongyu Li\textsuperscript{2}, Xuanqi Zeng\textsuperscript{1}, Laura Smith\textsuperscript{2}, Kyle Stachowicz\textsuperscript{2}, Dhruv Shah\textsuperscript{2}, Linzhu Yue\textsuperscript{1}, Zhitao Song\textsuperscript{1}, Weipeng Xia\textsuperscript{1}, Sergey Levine\textsuperscript{2}, Koushil Sreenath\textsuperscript{2},  Yun-hui Liu\textsuperscript{1}
\thanks{
\textsuperscript{1} The Chinese University of Hong Kong, China

\textsuperscript{2} University of California, Berkeley, USA

E-mail: hbzhang@mae.cuhk.edu.hk}
}



\maketitle

\begin{abstract}
The enhanced mobility brought by legged locomotion empowers quadrupedal robots to navigate through complex and unstructured environments. 
However, optimizing agile locomotion while accounting for the varying energy costs of traversing different terrains remains an open challenge.
Most previous work focuses on planning trajectories with traversability cost estimation based on human-labeled environmental features. 
However, this human-centric approach is insufficient because it does not account for the varying capabilities of the robot locomotion controllers over challenging terrains.
To address this, we develop a novel traversability estimator in a robot-centric manner, based on the value function of the robot's locomotion controller. 
This estimator is integrated into a new learning-based RGBD navigation framework.
The framework employs multiple training stages to develop a planner that guides the robot in avoiding obstacles and hard-to-traverse terrains while reaching its goals.
The training of the navigation planner is directly performed in the real world using a sample efficient reinforcement learning method that utilizes both online data and offline datasets.
Through extensive benchmarking, we demonstrate that the proposed framework achieves the best performance in accurate traversability cost estimation and efficient learning from multi-modal data (including the robot's color and depth vision, as well as proprioceptive feedback) for real-world training. Using the proposed method, a quadrupedal robot learns to perform traversability-aware navigation through trial and error in various real-world environments with challenging terrains that are difficult to classify using depth vision alone. 
Moreover, the robot demonstrates the ability to generalize the learned navigation skills to unseen scenarios. Video can be found at \url{https://youtu.be/RSqnIWZ1qks}.
\end{abstract}


\section{Introduction}

\begin{figure}[!t]
\centering
\includegraphics[width=3.4in]{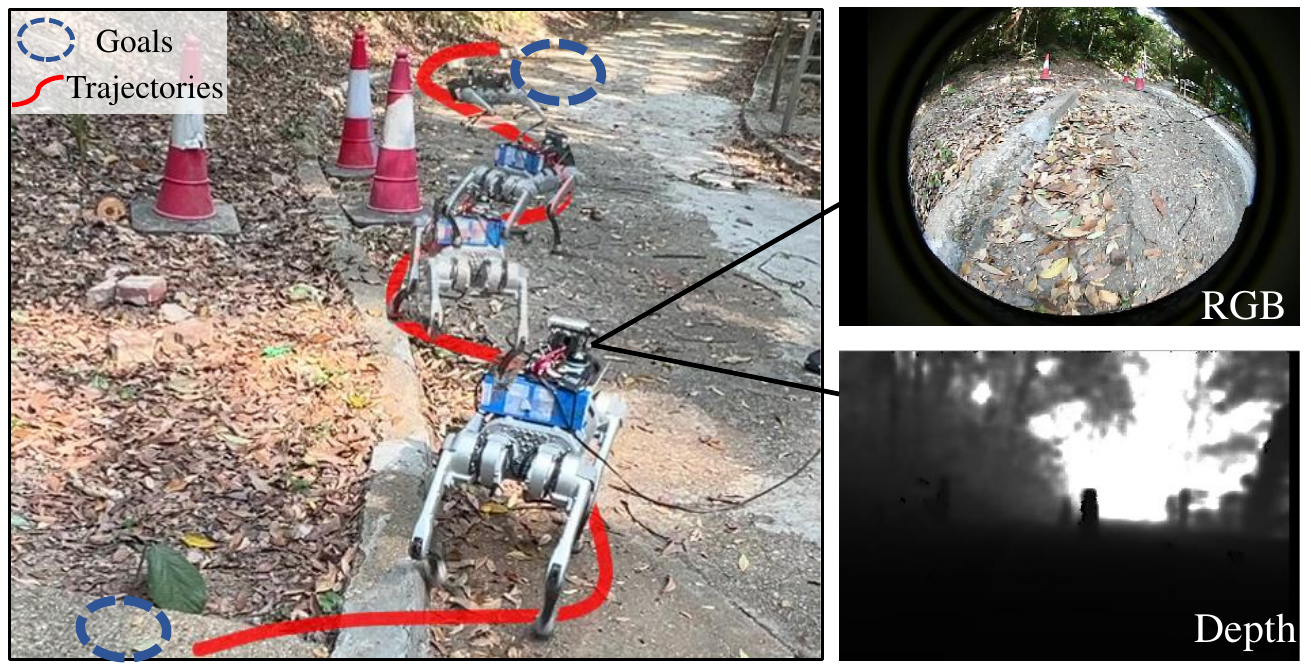}\caption{Our proposed framework enables a quadrupedal robot to learn to avoid hard-to-traverse terrains (such as
muddy areas covered by leaves) and obstacles from its own interactions through training directly in the real world. The robot learns to utilize its onboard color and depth vision sensors to
identify challenging terrains and obstacles while navigating toward the goals, marked by the blue dashed circles. The inclusion of RGB images allows the robot to identify additional terrain textures that are difficult to perceive with depth images alone during the navigation.}
\label{quadruped-trajectory}
\vspace{-0.5cm}
\end{figure}

With recent advancements in legged locomotion control~\cite{Zhuang2023RobotPL,lee2020learning,Perceptive-mpc}, quadrupedal robots can now perform robust locomotion over various terrains, including challenging ones like stairs or slippery surfaces.
When deploying these robots in outdoor environments, they need to identify whether a region is traversable by, for example, detecting obstacles and avoiding these areas accordingly.
The question is whether we can step beyond binary traversability estimation to include the cost of traversing different terrains, thus preferring paths with lower traversal costs.
For example, as shown in Fig.~\ref{quadruped-trajectory}, we prefer the robot to walk on a concrete walkway instead of a muddy off-road area, while still avoiding obstacles. 
It could be more optimal for the robot in terms of better stability of the locomotion controller, energy efficiency, and better maintenance of the hardware, \textit{i.e.}, having a low traversability cost.

However, estimating the traversability cost of terrain and planning to avoid high-cost regions for legged robots is a challenging problem. 
It requires the robot to interpret visual textures and corresponding physical properties, such as friction and elasticity, of different terrains.
For example, when humans encounter a muddy road with high traversability costs, we first see the mud's texture and infer its difficulty based on \emph{previous experience}.
We highlight the importance of learning from previous experience as it's difficult to accurately estimate the terrain's actual cost until we traverse it ourselves.
For legged robots to develop similar intelligence, they must address these multi-modal learning challenges, integrating data from various sources. 
To estimate terrain traversability cost, the robot needs to fuse color data (terrain texture), depth data (identifying non-traversable regions like obstacles), and proprioceptive feedback (implicitly encoding the physics) in real time and the real world.
This problem is further complicated by the complex dynamics of quadrupedal robots, which require not only a robust locomotion controller itself but also consider the stability of such a controller over different terrains.

To address this and achieve traversability-aware quadrupedal navigation autonomy in the wild, in this work, we propose developing a multi-stage hierarchical reinforcement learning (RL) framework. This framework learns from both simulation and real-world data including different modalities.
Our approach emphasizes the development of an online traversability estimator based on the value function of the RL-based low-level locomotion controller. 
The estimated traversability cost, which assesses control performance, is then used to train a high-level traversability-aware navigation planner that integrates both depth and color images.
We also hypothesize that training the planner directly in the real world frees us from relying on accurate and photo-realistic simulators, and removes the need to deal with sim-to-real gaps.

The central contribution of this work is the development of a hierarchical RGBD-based traversability-aware navigation-locomotion framework using reinforcement learning. This framework enables the robot to learn to reach goals, avoid obstacles, and minimize trajectory traversability cost from color and depth information in the real world within 15 minutes.
To quantitatively evaluate terrain traversability cost, we leverage the value function obtained during the training of the low-level locomotion controller using the actor-critic method. 
Traditionally, the critic is only used during training and not during deployment.
However, in this work, we deploy the trained value function to evaluate control performance across different terrains, thus representing the terrain traversability cost. 
This estimator, based solely on the robot's proprioceptive feedback, can also be used in different modules. 
To develop a planner that interprets terrain texture and geometry in the real world, we enhance an off-policy RL framework to learn from both online data and offline demonstrations. 
Validated through simulation and real-world experiments, our framework demonstrates better performance in terms of better estimation of the traversability, and improved generalization to new environments compared to baseline methods.

\section{Related work}

Most previous attempts divide the traversability-aware navigation problem into two subproblems: traversability estimation and path planning. 
\subsection{Traversability estimation}
To estimate the terrain traversability cost, some previous work directly extracts features from the environment. These methods usually define traversability by empirically judging the cost from meshes~\cite{brandao2020gaitmesh}, point clouds~\cite{Dixit2023STEPST}, or depth images~\cite{yang2023iplanner}.
Others try to classify the terrains from semantic categories \cite{schilling2017geometric,bradley2015scene,roth2023viplanner}, 
often mapping semantic features to traversability cost based on human-provided labels.
These methods typically rely on human prior knowledge rather than the robot's own experiences~\cite{stachowicz2023fastrlap}.
While these estimators are robot-agnostic and easily transferable across different embodiments, they fail to consider the robot's real interaction experiences. For example, some terrains may be difficult for lightweight robots but manageable for heavier ones, and vice versa.

To consider the robot's experience while traversing different terrains, previous work has suggested various metrics. The most straightforward approach is to consider control performance metrics such as tracking error of commands (e.g., velocity)~\cite{frey2023fast} or IMU variance\cite{chavez2018learning}. 
These quantitative metrics are used as labels for supervised learning of environment representations.
Other approaches extend this research by learning estimators to predict traversal time~\cite{margolis2023learning}, selected environment physics parameters~\cite{loquercio2023learning}, or collision and fall indicators~\cite{fu2022coupling} from a history of robot's proprioception feedback. While these methods link terrain traversability to specific control performance metrics, they do not capture the whole-body performance (including velocity tracking, base stability, motion smoothness, energy efficiency, and more) of a legged robot's full-order dynamics driven by its controller.
In this work, we address this by developing a traversability estimator from proprioception based on the value function of the RL-based locomotion controller.

\subsection{Learning based navigation}
While numerous methods can tackle the planning problem for quadrupedal robots~\cite{yasuda2020autonomous}, we focus on learning-based methods that provide a planning policy directly from the robot's vision input.
Some prior work opts to use supervised learning by imitating expert trajectories or demonstrations provided by humans, such as~\cite{loquercio2021learningflight, sadat2020perceive, shah2023vint}. 
In these methods, the robot is trained to reason expert actions from vision-based observations along demonstrated trajectories. 
This approach enables the robot to acquire behaviors quickly by mimicking successful strategies. 
However, the traversability cost can differ significantly across different embodiments. 
If the robot only learns from demonstrations provided by others (\emph{e.g.}, humans), it cannot update and optimize its planning strategy based on its own experiences: the planning policy is limited to offline data.

Reinforcement learning offers an approach that allows the robot to update its planning strategy based on its interaction with the environment, \emph{i.e.}, to learn from online data.
Usually, RL requires extensive data through trial and error, therefore, some prior attempts leverage photo-realistic simulation to train the robot extensively and directly transfer it to the real world, primarily in indoor environments such as~\cite{pmlr-v155-jain21a,sorokin2022learning,wijmans2019dd,kahn2021badgr}. 
However, for outdoor environments where large sim-to-real gaps exist, due to factors like varying lighting conditions and terrain types~\cite{frey2023fast}, zero-shot transfer of a color-vision-based policy from sim to real is still challenging.

Therefore, some other attempts try to train the planning policy directly with data collected in the real world, such as on racing cars with states-based data~\cite{rosolia2019learning} or vision-based data~\cite{stachowicz2023fastrlap}. 
For more complex robots, like quadrupedal robots, there have been attempts to train state-based control policies from scratch~\cite{smith2021legged,Wu22CoRL_DayDreamer}. 
However, using RL to deploy real-world training for quadrupedal robots to extract vision features, which are in a much higher dimension, remains an open question. 
This requires a sampling-efficient yet capable online learning algorithm to learn from high-dimension vision data.
To tackle this and obtain a traversability-aware planning policy for real-world deployment, we enable the quadrupedal robot to take advantage of both online data (its own experience) and offline data, such as expert demonstrations and embeddings from a pretrained color image encoder.

\section{Formulation of Traversability-Aware Legged Navigation and Its Framework}\label{Section:Problem}
In this section, we formulate the problem of traversability-aware navigation using quadrupedal robots by RL and develop the framework to tackle this problem.

\subsection{Problem Formulation}
\label{subsec:problem_formulation}
The navigation problem builds on the general robot navigation objective: learn to have a collision-free path towards the goal, but augmented with an extra task of minimizing the trajectory-wise traversability cost. To be specific, given a sequence of depth images \(I_{t:t-m}^{\text{depth}}\), a sequence of color images $I_{t:t-n}^{\text{rgb}}$ at time $t$ and a goal position \(\mathbf{x}^d\), the planner $\pi^p_{\text{rgbd}}$ should learn to generate velocity steering commands $\mathbf{a^p_t}$ for the locomotion controller $\pi^c$ to avoid obstacles and reach goal point \(\mathbf{x}^d\), while minimizing the trajectory-wise traversability cost (\textit{i.e.}, traversability-aware). To be specific, the entire problem can be divided into three subtasks:

\begin{itemize}[leftmargin=0.5cm]
\item \textbf{Goal Reaching}: The robot should reach the desired goal \(\mathbf{x}^d\) in minimal time. 
\item \textbf{Obstacle Avoidance}: The robot should avoid static and dynamic obstacles while traveling to the goal. Obstacle information can be perceived by the robot's depth vision.
\item \textbf{Traversability Cost Minimization}: The robot should be aware of the interplay between terrain traversability and locomotion capability, minimizing the traversability cost over trajectories. The terrain's traversability should depend on the robot's own interaction experiences instead of predefined labels, semantic classifications, or human-centric experiences.
\end{itemize}

\begin{figure*}[!t]
\centering
\includegraphics[width=0.99\linewidth]{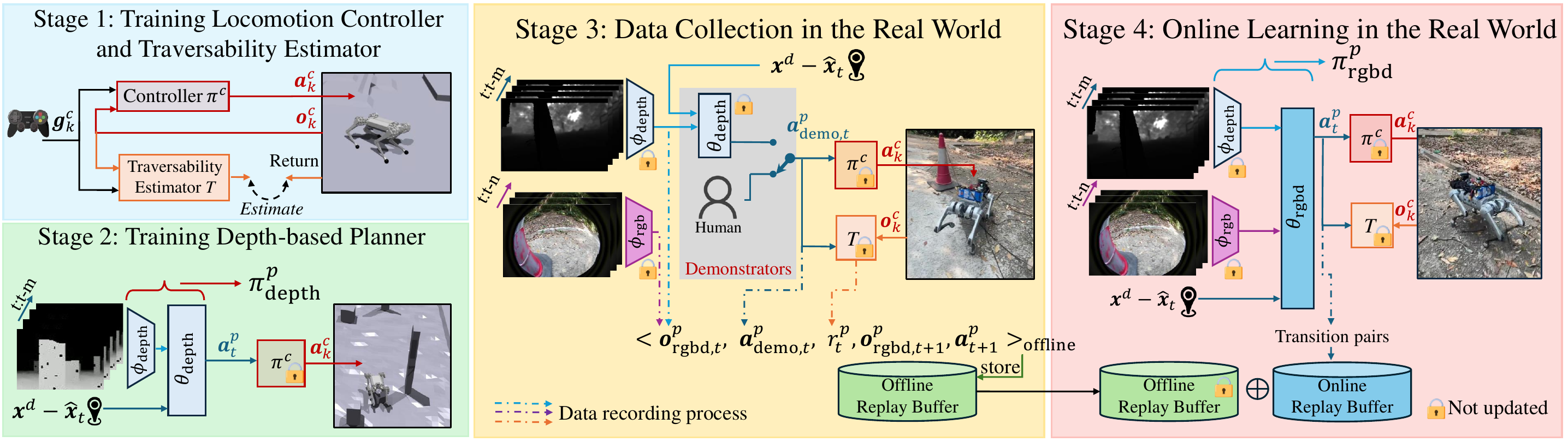}
\caption{The proposed training framework of high-level traversability-aware planner with RGBD input is divided into four stages. In Stage 1, the robust locomotion controller $\pi^c$ tracking desired velocity commands $\mathbf{g}^c_k$ and the corresponding traversability estimator $T$ are obtained as elaborated in Fig.~\ref{trv_est}. During Stage 2, a depth-based goal-reaching planner $\pi^p_{\text{depth}}$ is trained using RL with access to the depth input. This planner will provide demonstrations in the next stage.
In Stage 3, we collect real-world rollouts into datasets from both the depth-based planner \(\pi^p_{\text{depth}}\) and a human demonstrators.
Data are collected in the form of transition samples.
Both color and depth information are recorded. Actions \(\mathbf{a}^p_{\text{demo},t}\) are recorded from demonstrators, and the online estimated traversability cost \(T\) is included as part of the reward \(r^p_t\). Finally, during stage 4, the proposed planning policy $\pi^p_{\text{rgbd}}$ is trained using both the samples from the offline dataset and the newly collected transition pairs using RLPD~\cite{ball2023efficient}.}
\label{diagram}
\vspace{-0.5cm}
\end{figure*}

\subsection{Framework}
\label{subsec:frame}
The proposed traversability-aware hierarchical-legged navigation framework consists of two parts: (1) a low-level locomotion controller $\pi^c$ for the quadrupedal robot, which can track varying velocity and turning commands over different terrains, and (2) a high-level planner $\pi^p_{\text{rgbd}}$ that considers the navigation goal, surroundings, and terrain textures to specify control commands. As illustrated in Fig.~\ref{diagram}, the training of this framework consists of four stages.

\subsubsection{\textbf{Stage 1}: Training a Robust Locomotion Controller and a Traversability Estimator}
Building the framework from scratch, we first obtain a robust locomotion controller \(\pi^c\) for the quadrupedal robot using RL, as shown in Fig.~\ref{trv_est}. 
Our novel design includes the development of an estimator for overall control performance changes induced by changes in terrain traversability \(T\). 
This estimator is based on the critic network \(V^c\) obtained during the training of the control policy. 
Several design choices are implemented to ensure this estimator's functionality for traversability estimation on the real robot, detailed in Sec.~\ref{subsec:estimate}. Both the controller and estimator are trained in simulation.

\subsubsection{\textbf{Stage 2}: Training a Depth-Based Navigation Planner as a Demonstrator}
After obtaining the locomotion controller, we start training the planner \(\pi^p_{\text{depth}}\) on top of it. We aim to train a zero-shot deployable planner with only depth input in simulation, which will later act as a demonstrator during real-world data collection. The objective of this planner is to drive the robot to a target location while avoiding nearby obstacles, which are perceived through depth images. This planner will act as a demonstrator for autonomous data collection in the real world. Details will be introduced in Sec.~\ref{subsec:depth_based_planner}.

\subsubsection{\textbf{Stage 3}: Data Collection in the Real World}
To increase sample efficiency for the real-world training, we first collect rollouts in the real world with the trained depth-based planner \(\pi^p_{\text{depth}}\).
Additionally, our framework allows for the inclusion of human demonstrations, where human experts directly navigate the robot.
During these rollouts, data is collected in the form of transition pairs: \(\langle \mathbf{o}^p_{\text{rgbd},t}, \mathbf{a}^p_t, r^p_t, \mathbf{o}^p_{\text{rgbd},t+1},\mathbf{a}^p_{t+1} \rangle\). These rollouts allow us to collect extended observations $\mathbf{o}^p_{\text{rgbd}}$, including RGB information that provides additional visual texture features of terrains and surroundings. The reward \(r^p_t\) collected during rollouts includes the estimated traversability cost \(T\) based on the robot's real experience interacting with the terrains.
The details of this stage are discussed in Sec.~\ref{subsec:offline_data}.


\subsubsection{\textbf{Stage 4}: Online Learning in the Real World}
After collecting the offline dataset, we can train a new planning policy $\pi^p_{\text{rgbd}}$ from scratch by an off-policy RL method~\cite{pmlr-v80-haarnoja18b} using both the offline dataset \(\mathcal{D}_{\text{offline}}\) and an online replay buffer \(\mathcal{D}_{\text{online}}\), as illustrated in Fig.~\ref{diagram}. Rollouts from the current policy $\pi^p_{\text{rgbd}}$ are collected into the online replay buffer \(\mathcal{D}_{\text{online}}\) and maintain the same format as those collected in the offline dataset. During training, the robot explores different trajectories, encounters varying traversability costs, and is encouraged to avoid high-cost regions by interpreting the cost through vision inputs of terrain textures.
The details of training is introduced in Sec.~\ref{subsec:online_rl}.

\section{Locomotion Control and a Traversability Estimator}\label{sec:control_estimate}
In this section, we develop the first building block of the entire framework: the controller $\pi^c$ for robust quadrupedal locomotion and a corresponding traversability estimator $T$ as mentioned in the Stage 1 of the framework before.

\subsection{Legged Locomotion Control over Complex Terrains}\label{subsec:control}
We first leverage RL to develop a robust blind locomotion controller for the quadrupedal robot, enabling it to traverse different terrains while tracking varying velocities. 
We use an actor-critic algorithm, PPO~\cite{schulman2017proximal}, where the goal-conditioned actor policy $\pi^c$  and critic network \(V^c\) are trained jointly in simulation.
We adopt a symmetric actor critc structure to ensure that the critic can also be inferenced on the real-world robot. 
The controller is designed to track varying goal commands \(\mathbf{g}_k\), including walking velocities in the sagittal and lateral directions \(\dot{q}^d_{[x,y],k}\) and turning rate \(\dot{q}^d_{\psi,k}\).
The reward is designed to accomplish different control sub-objectives with different weights, including tracking given commands, following parameterized reference motion obtained by a central pattern generator~\cite{free-gait}, and several auxiliary terms including stabilizing base orientation, minimizing base vertical velocity, joint velocity, and acceleration. 
We randomize robot dynamics and terrain features and the training is done in Isaac Gym~\cite{liang2018gpu}.

\subsection{Estimation of Control Performance Based on Value Function}\label{subsec:estimate}
Our design originates from the following intuition: if the robot can traverse challenging terrains easily, the estimated traversability cost will still be low. This indicates that the level of traversability should be scaled according to the strength of the locomotion controller.
The critic network estimates the value, 
Interestingly, the traversability cost of the terrain can be inversely related to the estimated value from the critic network of locomotion representing the expected future accumulated reward. 

However, directly using the critic network for real-world evaluation of traversability is challenging. 
A frequent problem we encountered is that the estimated value is also sensitive to changes in the given commands \(\mathbf{g}_k\) using the goal-conditioned controller. 
For example, the critic return varies even when the robot locomotes on flat ground due to imperfect velocity tracking given different commands \(\mathbf{g}_k\).
This issue is further compounded by the sim-to-real gap: the critic may produce incorrect value predictions due to domain shifts, potentially overfitting to the simulation.

To address the issues above, we develop a traversability estimator \(T\) based on the critic network \(V^c\) trained with the actor-critic algorithm. Our major designs can be categorized into two key points:
\begin{itemize}
    \item We design a flat ground baseline \(V_{\text{flat}}^c(\mathbf{g}^c_k)\) and treat the value drops from the baseline to the critic value as the estimated cost. This aims to remove the dependency of the critic value on the given command \(\mathbf{g}_k\). 
    \item We explicitly calculate the uncertainty level $\sigma$ of the estimation from critic $T_{\text{value}}$ and use it to dynamically adjust the trust between the critic's estimation $T_{\text{value}}$ and the estimation from a backup estimation $T_{\text{track}}$. This approach enhances the reliability of the estimation when the prediction from the critic is uncertain.
\end{itemize}

As explained in Algorithm~\ref{alg:alg_trav}, it contains three steps to obtain the final estimated traversability cost $T$ to use in the real world, as detailed in the following. 

\begin{figure*}[!t]
\centering
\includegraphics[width=0.99\linewidth]{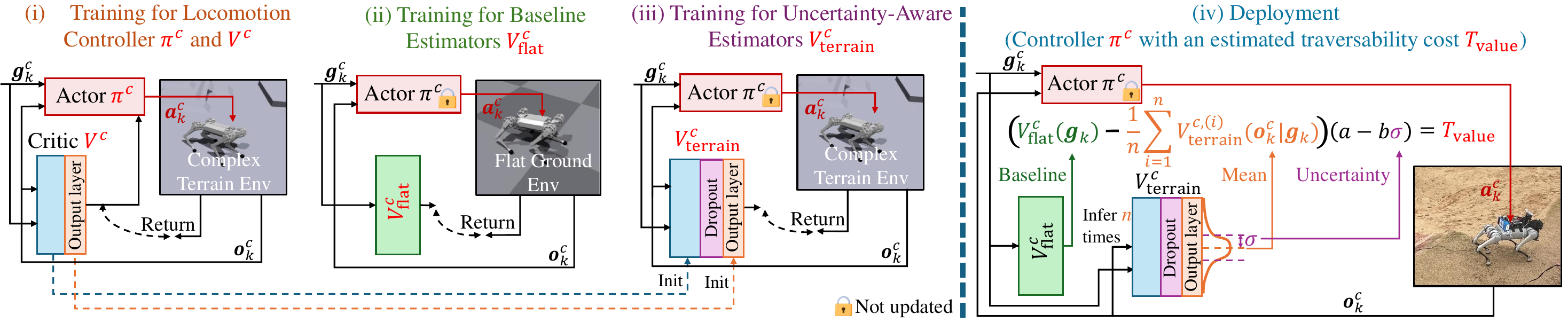}
\caption{The training framework for locomotion and traversability estimator. (i) The goal-conditioned controller (actor policy $\pi^c$) is trained to perform quadrupedal trotting gaits while tracking velocity commands $\mathbf{g}^c_k$ over various terrains. (ii) After obtaining $\pi^c$, a baseline value function $V^c_{\text{flat}}$ is trained to estimate the return while the robot is walking on flat ground using the same actor policy.
(iii) To create an uncertainty-aware estimator ($V^c_{\text{terrain}}$), we add a dropout layer before the output of $V^c$ and fine-tune it on complex terrains. 
(iv) In real-world deployment, we perform $n$ inferences using $V^c_{\text{terrain}}$ for the same goal $\mathbf{g}_k$ and observation $\mathbf{o}^c_k$. The standard deviation $\sigma$ of these inferences measures uncertainty. The mean value is adjusted by subtracting the baseline $V^c_{\text{flat}}$ to reduce bias induced by the given goal $\mathbf{g}^c_k$. The negative of this difference is then multiplied by an adaptive weight $(a - b\sigma)$, where $a$ and $b$ are tunable parameters, resulting in the traversability estimation $T_{\text{value}}$.}
\label{trv_est}
\vspace{-0.3cm}
\end{figure*}

\textbf{Training Baseline Value Estimator:}
As illustrated in Fig.~\ref{trv_est} (ii), we create a flat-ground value estimator \(V_{\text{flat}}^c(\mathbf{g}^c_k)\) to capture the effects of commands \(\mathbf{g}^c_k\) on performance. This estimator is trained by running the goal-conditioned locomotion controller \(\pi^c\) on flat ground, recording the ground truth return for various commands $\mathbf{g}^c_k$. The value estimator learns to predict the expected return from command input $\mathbf{g}^c_k$, providing a baseline to isolate the effects of terrain changes from command inputs.

\textbf{Training Uncertainty-Aware Value Estimator:}
As shown in Fig.~\ref{trv_est} (iii), we use the parameters of the pre-trained \(V^c\) and add a \emph{dropout} layer right before its output layer, initializing a value estimation network \(V_{\text{terrain}}^c\).

For inference, we perform \(n\) forward passes with the same observation (\(\mathbf{o}_k\)) and condition (\(\mathbf{g}_k\)). We then calculate the mean of the predictions \(\frac{1}{n} \sum_{i=1}^{n} V_{\text{terrain}}^{c,(i)}(\mathbf{o}_k|\mathbf{g}_k)\) as the estimated return, and the standard deviation \(\text{Var}(V_{\text{terrain}}^{c,(i=1:n)}(\mathbf{o}_k|\mathbf{g}_k))\) as the uncertainty metric, inspired by~\cite{Kahn2017UncertaintyAwareRL}.
This procedure acts like ensemble methods, where disagreement among multiple predictions indicates higher uncertainty, suggesting that the predicted return is less trustworthy and should be treated with caution.

\begin{remark}
The dropout layer is only added after the actor and critic networks are trained. The reason we do not concurrently train the critic with dropout is that the added dropout layer introduces a higher variance and fitting error, making the training for the actor network unstable.
\end{remark}

\begin{algorithm}[!htp]
\caption{Traversability Estimation}\label{alg:alg_trav}
\begin{algorithmic}
\STATE \textbf{GIVEN } \text{Trained locomotion controller and its critic network } $\pi^c$, $V^c$
\STATE \textbf{Output} $T_{\text{value}}$ with uncertainty
\STATE
\STATE // \textbf{Train} baseline value estimator $V^c_{\text{flat}}$
\STATE \textbf{for} $j=1:\text{max\_training\_iteration}$
\STATE \hspace{0.5cm} \textbf{for} $k=1:\text{batch\_size}$
\STATE \hspace{1cm} Sample $\mathbf{g}^c_k=\dot{q}^d_{[x,y,\psi],k}$ under \textbf{flat ground}
\STATE \hspace{1cm} Observe $\mathbf{o}^c_k$
\STATE \hspace{1cm} Run $\pi^{\text{c}}(\mathbf{o}^c_k|\mathbf{g}^c_k)$ and collect future return $R_k$
\STATE \hspace{0.5cm} \textbf{end for}
\STATE \hspace{0.5cm} Update $V^{c}_{\text{flat}}(\mathbf{g}^c_k)$ to fit return $R_k$
\STATE \textbf{end for}
\STATE 

\STATE // \textbf{Train} uncertain-aware estimator $V^c_{\text{terrain}}$
\STATE \textbf{Initialize}  Add a dropout layer next to the final layer of  $V^c$ and obtain initial $V^{c}_{\text{terrain}}$
\STATE \textbf{for} $j=1:\text{max\_training\_iteration}$
\STATE \hspace{0.5cm} \textbf{for} $k=1:\text{batch\_size}$
\STATE \hspace{1.0cm} Sample $\mathbf{g}^c_k=\dot{q}^d_{[x,y,\psi],k}$ and \textbf{complex terrain types}
\STATE \hspace{1.0cm} Observe $\mathbf{o}^c_k$
\STATE \hspace{1.0cm} Run $\pi^c(\mathbf{o}^c_k|\mathbf{g}^c_k)$ and collect future return $R_k$
\STATE \hspace{0.5cm} \textbf{end for}
\STATE \hspace{0.5cm} Update $V^{c}_{\text{terrain}}(\mathbf{o}^c_k|\mathbf{g}^c_k)$ with \emph{dropout} to fit return $R_k$
\STATE \textbf{end for}
\STATE 
\STATE // \textbf{Estimate} $T_{\text{value}}$ with uncertainty 
\STATE Observe $\mathbf{o}^c_k$, $\mathbf{g}^c_k$
\STATE \textbf{for} $i=1:n$ inference
\STATE \hspace{0.5cm} Run $i$-th forward pass on $V^{c,(i)}_{\text{terrain}}(\mathbf{o}^c_k|\mathbf{g}^c_k)$ with \emph{dropout}
\STATE $T_{\text{value}} \leftarrow$ $V^c_{\text{flat}}(\mathbf{g}^c_k) - \text{mean}(V^{c,(i=1:n)}_{\text{terrain}}(\mathbf{o}^c_k|\mathbf{g}^c_k))$ 
\STATE with \textbf{uncertainty} $\sigma(V^{c,(i=1:n)}_{\text{terrain}}(\mathbf{o}^c_k|\mathbf{g}^c_k))$
\STATE \textbf{end for}
\end{algorithmic}
\label{alg1}
\end{algorithm}

\textbf{Composition of Traversability Estimator:} 
The difference between the two expected returns: \(V_{\text{flat}}^c(\mathbf{g}^c_k) - V_{\text{terrain}}^c(\mathbf{o}^c_k|\mathbf{g}^c_k)\), indicates the changes in control performance induced by the changes in terrain, diminishing the effects brought by changes in conditions using the goal-conditioned controller. 
If the difference is smaller, it means the robot can traverse the current terrain with performance more similar to traversing flat ground using the same commands, thus the traversability cost $T_{\text{value}}$ based on value estimation is lower, and vice versa.
Furthermore, we will choose not to trust the estimation if the prediction is very unreliable, indicated by high prediction uncertainty \(\text{Var}(V_{\text{terrain}}^{c,(i=1:n)}(\mathbf{o}_k|\mathbf{g}_k))\). In such cases, we use another heuristic-based metric, such as command tracking errors, as an alternative for the traversability estimation. The composition of such a traversability estimation is given by:
\begin{subequations}
\label{trav_eq}
\begin{align}
T &= w_1 T_{\text{value}}+ (1-w_1) T_{\text{track}} \\
T_{\text{value}} &= V^c_{\text{flat}}(\mathbf{g}^c_k)-\text{Mean}(V^{c,(i=1:n)}_{\text{terrain}}(\mathbf{o}^c_k|\mathbf{g}^c_k)) \\
T_{\text{track}} &= \Vert \dot{q}^d_{[x,y],k} - \hat{\dot{q}}_{[x,y],k}\Vert_2 \\
w_1 &= a - b \text{Var}(V^{c,(i=1:n)}_{\text{terrain}}(\mathbf{o}^c_k|\mathbf{g}^c_k))
\end{align}
\end{subequations}\noindent
where the weight \(w_1\) is inversely proportional (by tuneable gains $a,b>0$) to the uncertainty of the value estimation: lower uncertainty in the value estimation results in a higher weight for the value estimation $T_{\text{value}}$ in the overall traversability estimation $T$, as we trust this estimation more. Conversely, higher uncertainty will lead the overall traversability estimation to rely more on heuristic-based metrics such as command tracking $T_{\text{track}}$.

\begin{remark}
This method of evaluating the difference between a baseline return estimation and an estimator for changes in the environment is not limited to traversability estimation as focused in this work. It can serve as a general evaluator of RL-based control performance. For example, it can also assess the sim-to-real gap: by comparing a baseline estimator for nominal dynamics parameters with an estimator for different dynamics parameters under the same given goal, a significant gap indicates a large sim-to-real discrepancy.
\end{remark}

\section{Combing Depth and Color Vision for Traversability-Aware Navigation}
After obtaining low-level locomotion control \(\pi^c\) and a traversability cost estimator \(T\) that can be deployed on the real-world robot, we focus on the high-level goal-reaching planner in this section. 

\subsection{Depth-based Navigation Planner as a Demonstrator}\label{subsec:depth_based_planner}
As illustrated in Fig.~\ref{diagram}, we first train a planner \(\pi^p_{\text{depth}}\) with access to a depth camera to navigate the robot to a given goal \(\mathbf{x}^d\) in 2D Cartesian space while avoiding nearby obstacles. The trained planner will be zero-shot transferred to the real world and serve as a demonstrator in the next stage. 

This planner is also trained using RL, and its POMDP design is detailed as follows.
\subsubsection{Action}

The depth-based planner \(\pi^p_{\text{depth}}\) outputs the change in control commands \(\mathbf{a}^p_t = \mathbf{c}_t - \mathbf{c}_{t-1}\) with clamping to ensure smooth, bounded actions. This eliminates the need for reward tuning for smoothness. The planner is designed to operate at 5 Hz.

\subsubsection{Observation}

The observation of the planner \(\mathbf{o}^p_{\text{depth},t}\) consists of: (1) a sequence of depth images \(I_{t:t-m}^{\text{depth}}\), with a length of \(m+1\); (2) the robot's current velocity \(\hat{\dot{q}}_{[x,y,\psi],t}\) and the previous action \(\mathbf{a}^p_{t-1}\); and (3) the position vector from the robot to the goal \(\mathbf{x}^d - \hat{\mathbf{x}}_t\). This provides a short memory of the surroundings, an I/O feedback, and the direction towards the goal.

\subsubsection{Reward}
The reward for training this depth-based planner is defined as the weighted sum of three terms: \(r_{\text{goal}}\), \(r_{\text{FoV}}\), and \(r_{\text{collision}}\), each defined as follows:
\begin{subequations}
\begin{align}
\label{rewards-2a}
r_{\text{goal}} &= \hat{q}_{[x,y],t} \cdot (\mathbf{x}^d - \hat{\mathbf{x}}_t) + 10G_{\text{arrive}} \\
\label{rewards-2b}
r_{\text{FoV}} &= v_{x} \\
\label{rewards-2c}
r_{\text{collision}} &= -G_{\text{collision}}
\end{align}
\label{rewards}
\end{subequations}\noindent 

The first reward term, \(r_{\text{goal}}\), is the dot product of the robot's planar velocity \(\hat{q}_{[x,y],t}\) and the vector to the goal \(\mathbf{x}^d - \hat{\mathbf{x}}_t\), encouraging a movement toward the goal. \(G_{\text{arrive}}\) provides a binary reward upon arrival. \(r_{\text{FoV}}\) discourages backward walking to maintain a forward-facing Field of View (FoV), while \(r_{\text{collision}}\) penalizes the collisions.

\subsubsection{Policy Representation and Training Details}

The depth-based planner \(\pi^p_{\text{depth}}\) is a deep neural network with two components: a depth image encoder \(\phi_{\text{depth}}\) and a base MLP. 

We train the depth-based planner \(\pi^p_{\text{depth}}\) using PPO in Isaac Gym, where random obstacles (cylinders, boxes) and simulated noise are added to the depth images.

\subsection{Data Collection in the Real World}\label{subsec:offline_data}

To accelerate the subsequent online training in the real world, we choose to collect a real-world offline dataset with demonstration rollouts before the training. 
As illustrated in Fig.~\ref{diagram}, the planner used during rollouts comes from (1) the depth-based planner \(\pi^p_{\text{depth}}\) trained in simulation and (2) human demonstrations where a human operator teleoperates the robot to navigate to the goal.

While the robot navigates by demonstrators, we collect transition samples from the planner over the trajectories. The transition sample at timestep \(t\) is denoted as \(\langle \mathbf{o}^p_{\text{rgbd},t}, \mathbf{a}^p_t, r^p_t, \mathbf{o}^p_{\text{rgbd},t+1},\mathbf{a}^p_{t+1} \rangle\), and will be stored in the offline replay buffer $\mathcal{D}_{\text{offline}}$. 
During deployment, a fisheye camera is used to capture wide-range color images, which are encoded using a pre-trained color image encoder \(\phi_{\text{rgb}}\) trained on a large-scale dataset~\cite{shah2021rapid}. 
The details of the collected data are introduced as follows.

\subsubsection{Action}
The action \(\mathbf{a}^p_t\) during the rollouts is still the change in the given control commands \(\mathbf{a}^p_t = \mathbf{c}_t - \mathbf{c}_{t-1}\). Actions come from either the depth-based planner or human demonstrations.

\subsubsection{Observation}
The observation \(\mathbf{o}^p_{\text{rgbd},t}\) at timestep \(t\) includes the latent embeddings from both the pre-trained depth image encoder \(\phi_{\text{depth}}\) and color image encoder \(\phi_{\text{rgb}}\), the robot's estimated base velocity \(\hat{\dot{q}}_{[x,y,\psi],t}\), the previous planner action \(\mathbf{a}^p_{t-1}\), and the distance to the goal location \(\mathbf{x}^d - \hat{\mathbf{x}}_t\).

\subsubsection{Reward}
To emphasize the traversability awareness, we add a new term \(r_{\text{terrain}}\). The terrain reward is the negative of the estimated traversability cost \(T\) developed in Eq.~\eqref{trav_eq}. Also, we retain the goal-reaching term \(r_{\text{goal}}\)~\eqref{rewards-2a} and forward term \(r_{\text{FoV}}\)~\eqref{rewards-2b}, drop the collision reward (as we lack a collision detector onboard).
\begin{subequations}
\begin{align}
r^p_t &= 0.2 r_{\text{goal}} + 0.02 r_{\text{FoV}} + 0.5 r_{\text{terrain}} \\
r_{\text{terrain}} &= -T
\end{align}
\label{rewards2}
\end{subequations}\noindent

\subsection{Online Learning for Traversability-Aware Navigation}\label{subsec:online_rl}
After collecting the offline replay buffer, we begin training the final policy, the traversability-aware planner $\pi^p_{\text{rgbd}}$ that fuses depth and color images, and robot state feedback. 
This training is from scratch and is directly deployed on the robot hardware while exploring different terrains during navigation.

We retain the depth image encoder \(\phi_{\text{depth}}\) and color image encoder \(\phi_{\text{rgb}}\) and online train the base MLP. 
This approach avoids learning directly from raw image data, which is of much higher dimensionality, thus accelerating real-world training. 
The training objective is to enable the base MLP to properly utilize the latent embeddings from the depth and color images, along with the robot feedback, to avoid terrains with high traversability cost and obstacles while navigating to the goal.

The elements in the POMDP (such as observation, reward, and action) during this stage remain the same as those in the offline data collection stage, as described in Sec.~\ref{subsec:offline_data}. 
The key difference is that we now start updating the planner policy (which produces the action \(\mathbf{a}^p_t\)) using the data collected with the new policy. 
We adopt DroQ algorithm~\cite{hiraoka2022dropout} (modified SAC~\cite{pmlr-v80-haarnoja18b} with dropout in Q-value network), augmented it with RLPD.
During each policy update, the sampled batch consists of two parts: half of the data is sampled from the offline dataset \(\mathcal{D}_{\text{offline}}\) previously collected, and the other half is sampled from the online data \(\mathcal{D}_{\text{online}}\) produced by the current policy.
The update rule follows the standard SAC algorithm.
By leveraging the offline dataset, we can accelerate online training and reduce the need for extensive exploration during online learning.

\section{Simulation Validation}
\label{simulation_validation}

In this section, we instantiate our proposed framework in a high-fidelity simulator to analyze various design choices and benchmark against prior work. This allows us to conduct large-scale experiments and derive statistically significant insights with scalable and reliable simulation evaluations, which will direct the real-world deployment of our system.

\subsection{Simulation Setup}
\begin{figure}[!t]
\centering
\includegraphics[width=3.4in]{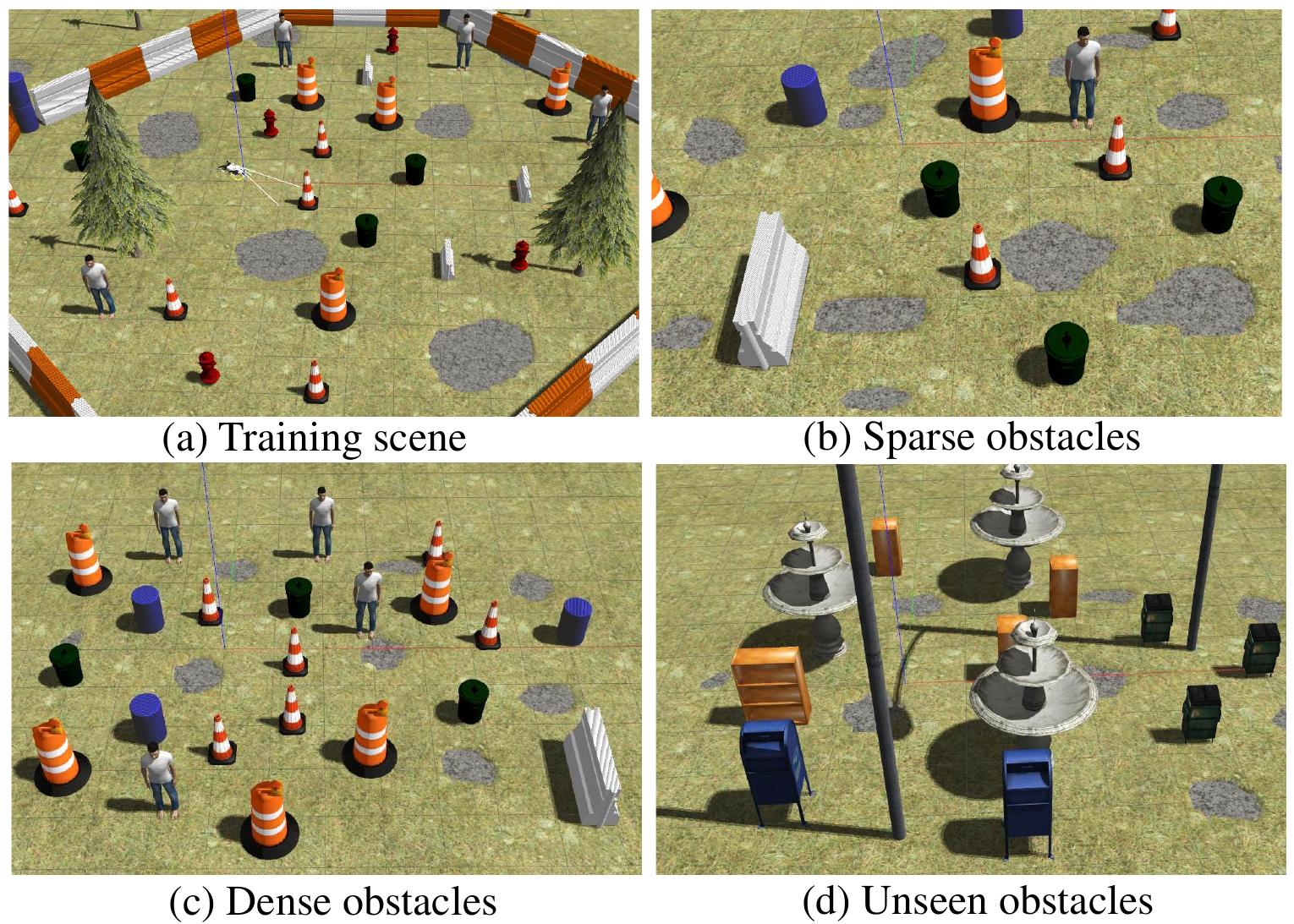}
\caption{The training scene (a) and three different testing scenes (b)(c)(d) in GAZEBO simulator are shown. To create environments with varying traversability, lower frictions are assigned to the areas of grey colors representing the water pool while nominal frictions are assigned to the green grasslands. Besides, various obstacles are randomly distributed inside. Policies are trained in the training scenario (a) and evaluated in testing scenarios (b), (c) and (d).}
\label{sim-env-pic-v2}
\vspace{-0.5cm}
\end{figure}

We developed a custom Gazebo simulation~\cite{koenig2004design}, as shown in Fig.~\ref{sim-env-pic-v2}, to study the various design choices. This simulation provides a consistent, controllable environment with physics simulation for the quadrupedal robot, including simulated depth and color ego-vision. 
In our setup, we assign \textbf{different friction coefficients} to different terrain types, such as lower friction in water pool regions and nominal values on grasslands, representative of real-world traversability. We also instantiate visually diverse obstacles, such as cones, trees, and even simulated humans, visualized in Fig.~\ref{sim-env-pic-v2}.

We deploy offline data collection and online training in Fig.~\ref{sim-env-pic-v2}(a) and test the trained framework in environments with \emph{sparse obstacles} (Fig.~\ref{sim-env-pic-v2}(b)), \emph{dense obstacles} (Fig.~\ref{sim-env-pic-v2}(c)), and \emph{unseen obstacles} (Fig.~\ref{sim-env-pic-v2}(d)). 

We report three metrics: average speed, traversability cost, and number of collisions, to quantitatively evaluate performance in tackling the navigation problem studied in this work, as described in Sec.~\ref{subsec:problem_formulation}. 
The simulator provides access to ground-truth traversability costs by examining the average time the robot walks in the water pool area (with lower friction). In each environment, 20 navigation goal points are randomly sampled in the environment.

\subsection{Estimating Traversability}
As this work focuses on traversability-aware navigation, reliably estimating the traversability cost is one of our main objectives. 
In this benchmark, we compare different methods to estimate the traversability cost while keeping all other components in the framework the same:
\begin{itemize}[leftmargin=0.5cm]
    \item \textbf{Ours}: The proposed traversability estimation combines the value estimate \(T_\text{value}\) from the locomotion controller and a backup metric, the velocity tracking error \(T_\text{track}\), as formulated in \eqref{trav_eq}.
    \item \textbf{$T$ from Tracking}: Only the velocity tracking error \(T_\text{track}\) is used to represent the traversability cost that the robot is currently traversing, as used in \cite{frey2023fast}.
    \item \textbf{$T$ from SysID}: Similar to~\cite{loquercio2023learning}, we adopt a teacher-student training paradigm. The student policy (resulting controller) features an encoder that uses the robot's long I/O history to identify preselected environment parameters input to the teacher policy. Here, we choose foot friction as the parameter to estimate and use the identified friction to evaluate the traversability cost (in simulation, we only change the ground friction for different terrain types).
\end{itemize}

\begin{figure}[!t]
\centering
\includegraphics[width=3.4in]{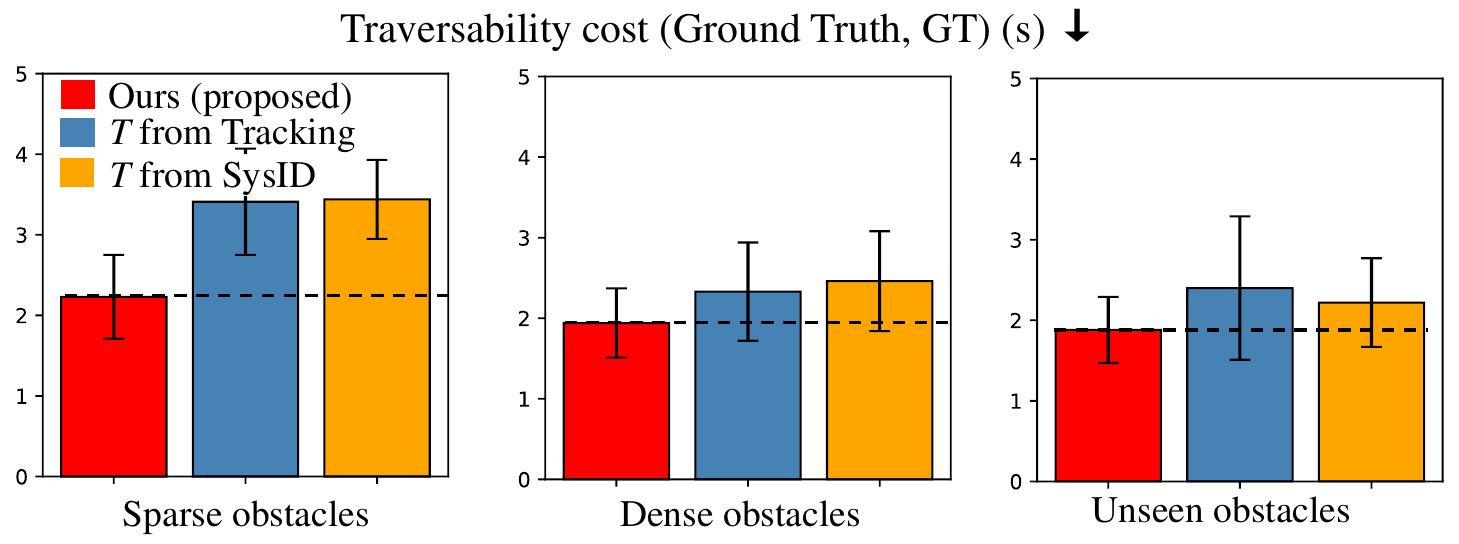}
\caption{Comparison between different choices of traversability estimation against ours is shown. The proposed traversability estimation method achieves at least \textbf{an 18\% reduction} in traversability cost across the three testing scenes compared to the other two baseline methods. The evaluation is done on three testing scenarios with ground truth traversability cost represented by the time of the robot walking in water pool areas, obtained directly from the simulator.}
\label{trv_metrics}
\vspace{-0.4cm}
\end{figure}

As shown in Fig.~\ref{trv_metrics}, our method, which includes the value estimate \(T_\text{value}\) to evaluate overall control performance, performs significantly better than using the velocity tracking error alone (\textbf{$T$ from Tracking}), resulting in a lower total traversability cost over the trajectory (\textit{i.e.} in \emph{sparse obstacles}: \(2.23\pm0.52\) s) compared with \textbf{$T$ from Tracking} (\(3.14\pm0.66\) s), which uses the velocity tracking error alone.

The results indicate that relying solely on heuristic-based metrics, such as velocity tracking error, fails to account for other factors that may not affect tracking performance but cause other degradations, such as foot slippage or body instability. These factors can be considered by the proposed estimator based on the controller's value function. 

Furthermore, compared with \textbf{$T$ from SysID}, which only identifies ground friction, our method also results in better trajectories with smaller accumulated traversability costs (ours $2.23\pm0.52$ s versus $3.44\pm0.49$ s). 
This improvement is due to the potential noise and unreliability in system identification.
Additionally, there are many factors in real-world terrain that cannot be parameterized, such as ground filled with cobblestones or sand, where the system identification method can potentially fail to distinguish their difference using our proposed method. 

\subsection{Importance of Using Both Depth and Color Vision}
We further highlight the importance of using both depth and color vision as input to the traversability-aware planner. In this benchmark, we develop different planners with different vision sources:
\begin{itemize}[leftmargin=0.5cm]
    \item \textbf{Ours}: The proposed method uses both depth vision and color vision, thus, our policy \(\pi^p_{\text{rgbd}}\) is based on RGBD data.
    \item \textbf{Depth Only}: A planner that only has access to depth vision, using the same length of past depth images as the proposed method.
    \item \textbf{Color Only}: A planner that only has access to color vision, using the same length of past color images as the proposed method.
\end{itemize}

\begin{figure}[!t]
\centering
\includegraphics[width=3.4in]{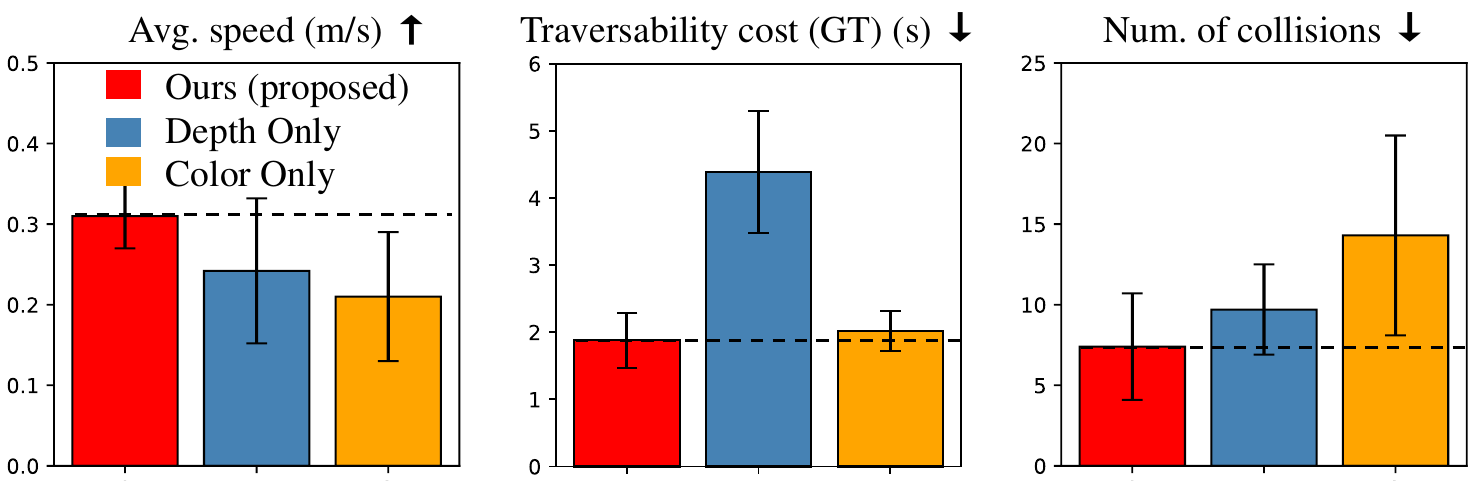}
\caption{Comparison between using single color or single depth vision as input against ours (using both color and depth) over the testing environment \textit{unseen obstacles} (Fig.~\ref{sim-env-pic-v2}(d)) is shown.The proposed method shows significantly better performance, especially in reducing traversability cost, outperforming the depth-only approach by \textbf{133\%}. Additionally, the proposed method reduces collisions with unseen obstacles during testing by \textbf{93\%} compared to the RGB-only approach. 
}
\vspace{-0.5cm}
\label{sensor_metrics}

\end{figure}

As shown in Fig.~\ref{sensor_metrics}, the \textbf{Depth Only} method results in the highest traversability cost over the trajectory. This is expected, as terrain texture, which reflects traversability, cannot be easily distinguished by depth vision alone, which only provides geometric information.

When comparing our method, which combines both depth and color vision, with the \textbf{Color Only} method, the performance in terms of obstacle avoidance is similar in \emph{sparse obstacles} (Fig.~\ref{sim-env-pic-v2}(b)) and \emph{dense obstacles} (Fig.~\ref{sim-env-pic-v2}(c)), environments similar to the training environment. However, the performance of \textbf{Color Only} drops significantly in \emph{unseen obstacles} (Fig.~\ref{sim-env-pic-v2}(d)), while our method still shows good performance. 
This comparison indicates that using only color images has a poor generalization ability over unseen obstacles. The inclusion of depth vision in our method alleviates this by providing geometric features.

\subsection{Different Sources of Offline Data}
We also note that the source of the offline data from which we collected also influenced the performance of the trained planner. In this benchmark, we validate the necessity of our choice on generating the offline data from a mixture of human demonstration and the demonstration from the robot itself (the depth-based planner).
\begin{itemize}[leftmargin=0.5cm]
    \item \textbf{Ours}: The proposed method uses both the rollouts from trained depth-based planner $\pi^p_{\text{depth}}$ and human demonstrations to generate offline data.
    \item \textbf{Human Demo. Only}: The rollouts are only collected from a human demonstrator to generate offline data.
    \item \textbf{Depth Demo. Only}: The rollouts are only collected from the trained depth-based planner $\pi^p_{\text{depth}}$ to generate offline data.
\end{itemize}

\begin{figure}[!t]
\centering
\includegraphics[width=3.4in]{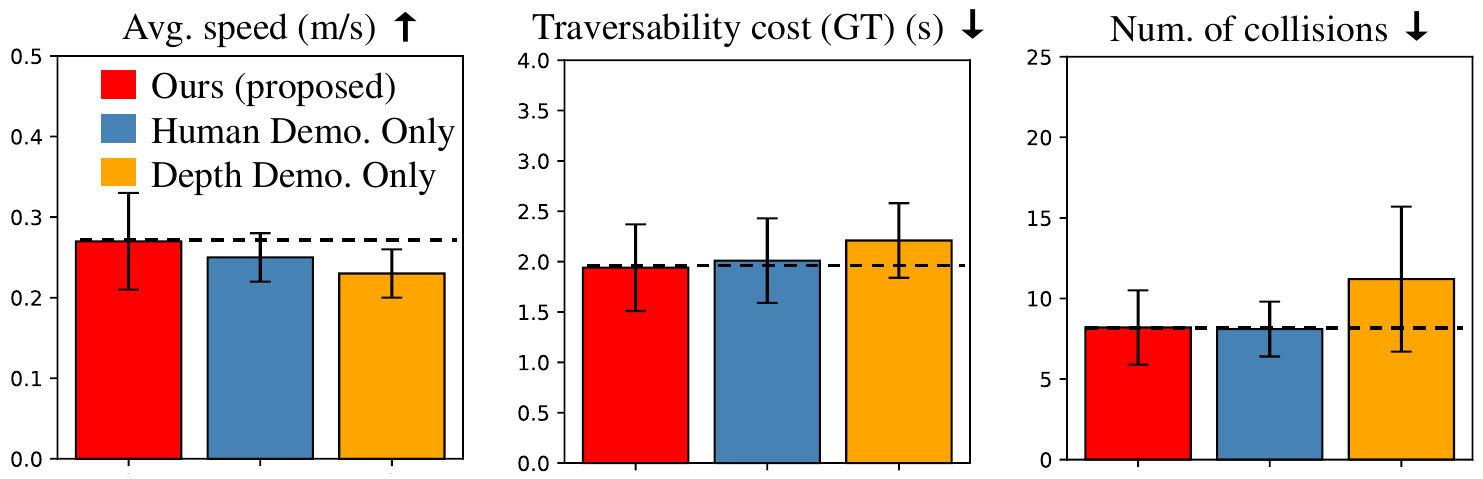}
\caption{Comparison between using a single data source of the offline data against ours (using demonstrations from both human experts and depth-based policy) is shown. The proposed method using the offline replay buffer containing the mixed demonstrations from both a human expert and a depth-based policy outperforms using a single data source of the offline dataset over environment \textit{sparse obstacles} (Fig.~\ref{sim-env-pic-v2}(b)). 
}
\label{source_metrics}
\vspace{-0.3cm}
\end{figure}
As shown in Fig.~\ref{source_metrics}, our method that uses the offline replay buffer which contains the mixed demonstrations from both a human expert and a pre-trained depth-based policy $\pi^p_{\text{depth}}$ outperforms the choice of using a single data source of the offline dataset on the overall performances. (\textbf{Human Demo. Only} and \textbf{Depth Demo. Only}). This is because a mixture of sources of the offline data enhances the data diversity and therefore helps to avoid potential overfitting and does better exploration. 

Compared with data collected solely from the depth-based planner (\textbf{Depth Demo. Only}), adding demonstrations from human experts provides a different distribution of the state-action pair to the offline dataset. 
This not only broadens the distribution of the collected trajectories but also introduces better trajectories in terms of minimizing traversability cost and obstacle avoidance, which are provided by human demonstrators.

However, if we only rely on the data collected by the human demonstrators, we also find there is performance degradation, especially for the goal-reaching objective (\textbf{Ours} \(0.27\pm0.06\) m/s versus \textbf{Human Demo. Only} \(0.23\pm0.03\) m/s). 
This is because human demonstrators may prefer the strategy that favors obstacle avoidance and lower traversability cost, but could be hard to consider minimizing traveling time. 

The data collected by the depth-based planner can overcome this problem as it explicitly handles the goal vector information with a well-trained planning policy.

\subsection{Learning from Both Offline and Online Data}
We compare our proposed learning method for the final planning policy with several benchmarks that share similar ideas of utilizing pre-collected offline data to benefit the training process. 
\begin{itemize}[leftmargin=0.5cm]
    \item \textbf{Ours}: We use a standard SAC augmented with RLPD~\cite{ball2023efficient}, which means that half of the update batch is sampled from the offline dataset $\mathcal{D}_{\text{offline}}$ and half of the update batch is sampled from the online dataset $\mathcal{D}_{\text{online}}$.
    \item \textbf{Vanilla SAC}: We train the planner policy using standard SAC~\cite{pmlr-v80-haarnoja18b} without the offline data.
    \item \textbf{Behavior Cloning (BC)}: We directly apply behavior cloning over the collected offline dataset $\mathcal{D}_{\text{offline}}$. A 2-norm action Mean Square Error (MSE) $||\mathbf{a}^{p}_{\text{demo}} - \pi^p_{\text{bc}}(\mathbf{o}^p_t)||_2$ is used as the loss function to minimize following the general behavior cloning setting, where \(\pi^p_{\text{bc}}\) is the target policy.
    \item \textbf{SAC with BC Loss}: Following~\cite{wu2023human}, we use a standard SAC augmented with an extra behavior cloning loss term added to the actor loss.    
\end{itemize}

Please note that all of offline data, excepting the one of the \textbf{BC}, is collected in the form of state-action transition pairs:  \(\langle \mathbf{o}^p_t, \mathbf{a}^p_{\text{demo},t}, r^p_t, \mathbf{o}^p_{t+1},\mathbf{a}^p_{\text{demo},t+1} \rangle\), while we only collect state-action pairs \(\langle \mathbf{o}^p_t, \mathbf{a}^p_{\text{demo},t} \rangle\) for \textbf{BC}.

\begin{figure}[!t]
\centering
\includegraphics[width=3.4in]{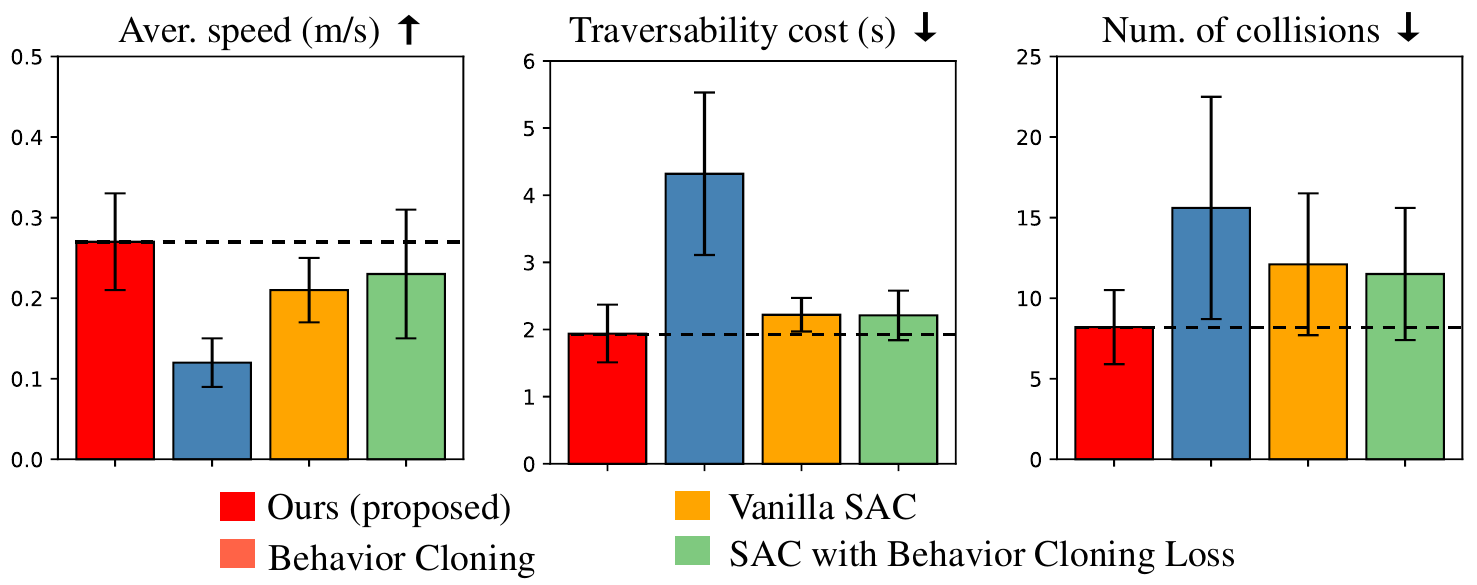}
\caption{Comparison of ways to incorporate offline dataset into the training process on testing scene \textit{dense obstacles} (Fig.~\ref{sim-env-pic-v2}(c)) is shown. The proposed method, which combines offline and online data using RLPD, demonstrates at least \textbf{14\%, 12\%, and 40\% better performance} across the three metrics. We validate that incorporating a pre-collected offline dataset is necessary concluding from the comparison to the benchmark \textbf{Vanilla SAC}. Then we compare with approaches that incorporate an offline dataset with supervised learning: \textbf{BC} and its variation \textbf{SAC with BC Loss}. 
}
\label{offline_metrics}
\vspace{-0.4cm}
\end{figure}

As shown in Fig.~\ref{offline_metrics}, our method, which uses RLPD to incorporate offline datasets into the training, significantly outperforms the other baseline methods.

Compared to the \textbf{Vanilla SAC} method, which does not use any pre-collected dataset, we find that incorporating an offline replay buffer, as in our proposed method, significantly improves overall training performance. This advantage is particularly evident in our case, where we are addressing multi-modal learning from high-dimensional visual data. The pre-collected offline replay buffer, $\mathcal{D}_{\text{offline}}$, provides a high-quality dataset that helps to warm up the learning process. This not only enhances the sampling efficiency for online learning but also broadens the coverage of the distribution of collected trajectories, thereby improving the generalization of the learned planner.

We also compared our method with another approach that incorporates a pre-collected dataset during learning, but without using reinforcement learning: \textbf{Behavior Cloning (BC)}. As shown in Fig.~\ref{offline_metrics}, \textbf{BC} demonstrates worse performance because it relies solely on regression from state-action demonstration pairs without any reward labels. This method is more susceptible to suboptimal demonstrations and fails to explore behaviors that could be more optimal than those demonstrated. 

In contrast, our proposed method is less sensitive to the quality of the demonstrations because the transition pairs collected by our method include a reward term. After online exploration and learning, the agent optimizes for accumulated rewards and surpasses the performance of suboptimal demonstrations.

We also established a benchmark using a variation of behavior cloning, specifically \textbf{SAC with BC Loss}, as proposed in~\cite{wu2023human}. This method augments standard RL with a behavior cloning term, providing an alternative way to incorporate offline demonstration data into an RL training pipeline. While this approach outperforms other baselines, as shown in Fig.~\ref{offline_metrics}, our proposed method still delivers superior results. The reason is that the inclusion of a behavior cloning term during policy optimization can restrict the agent's ability to explore more optimal solutions based on its own experiences. 

\subsection{Summary}
In summary, we justify several key design choices for training the RGBD-based planner policy and provide a comprehensive quantitative analysis. The findings are summarized in four key points:
\begin{itemize}[leftmargin=0.5cm]
    \item \textbf{Traversability Estimator}: We emphasize the use of value estimation $T_{\text{value}}$ from the locomotion controller. 
    This approach results in a more reliable and robust robot-centric traversability estimate.
    \item \textbf{Types of Vision Input}: We recommend using both color and depth information to enhance scene understanding and generalization of the outdoor environments, which benefits obstacle avoidance, terrain texture identification, and ultimately, traversability-aware navigation.
    \item \textbf{Sources of Offline Data}: We underscore the importance of collecting offline data from diverse sources. By incorporating both human-centric and robot-centric demonstrations, the dataset becomes more varied and beneficial for subsequent online learning.
    \item \textbf{Learning from Offline and Online Data}: We demonstrate that our proposed method, which integrates both offline and online data through reinforcement learning, improves sample efficiency and mitigates the effects of suboptimal demonstrations.
\end{itemize}

\section{Real World Experiments}
After substantiating the design choices underlying our framework and demonstrating that our method achieves the best performance in simulation, we now deploy the entire system in the real world to validate its effectiveness.

\subsection{Experiment Setup}
The experiments are conducted on a quadrupedal robot (Unitree Go1). Visual sensing is achieved using a fisheye camera for color images and a RealSense D400 for depth perception. Additionally, a 3D LiDAR is utilized for localization via LiDAR-inertial odometry~\cite{xu2021fast}, providing the robot's relative position to the target point in outdoor environments.

To validate the proposed methods, we selected three distinct real-world scenarios, as illustrated in Fig.~\ref{real_ps_robts}. These scenarios were chosen for their complex terrains, which include areas that are challenging to classify based solely on geometric information (\emph{e.g.}, depth vision).

\begin{itemize}[leftmargin=0.5cm]
    \item \textbf{Roads with Leaves} (Fig.~\ref{real_ps_robts}(a)): This scenario features terrain divided into difficult-to-traverse areas with piles of leaves and normal flat areas that are easier for quadrupeds to navigate. Construction cones are randomly scattered throughout as obstacles.
    \item \textbf{Cobblestone Roads} (Fig.~\ref{real_ps_robts}(b)): The terrain in this scenario consists of a mix of cobblestone and flat road areas. The cobblestone sections cause slipperiness, leading to unstable movements for the legged robot, making these areas difficult to traverse. A fixed streetlight also serves as an obstacle.
    \item \textbf{Bunker Roads} (Fig.~\ref{real_ps_robts}(c)): This scenario includes a sand-filled bunker in the middle, surrounded by flat grassy areas. The sand bunker is considered a difficult-to-traverse area as the quadruped often gets stuck. Construction cones are randomly placed as obstacles throughout the scenario.
\end{itemize}
In each scenario, two or three goal points are selected, with arrival defined as the robot's position being within a 0.5-meter radius of the goal point. The goal points are switched consecutively after the robot successfully reaches each one, allowing for continuous data collection and training.

For real-world implementation, we first collect real interaction data and store them in the offline replay buffer $\mathcal{D}_{\text{offline}}$, collecting approximately 2500 sample points through demonstrations from both a human expert and a trained depth-based planner $\pi^p_{\text{depth}}$. 
This process typically takes around 10 minutes. 
Subsequently, online learning using the collected offline data by RLPD is conducted to develop the final RGBD-based planner $\pi^p_{\text{rgbd}}$, as described in \textit{Stage 4} (Sec.~\ref{subsec:online_rl}).

\subsection{Traversability Estimation in the Wild}
As a critical component of the navigation framework, we first present the results of traversability estimation, highlighting the role of explicit uncertainty evaluation in real-world environments.

\begin{figure}[!t]
\centering
\includegraphics[width=3.3in]{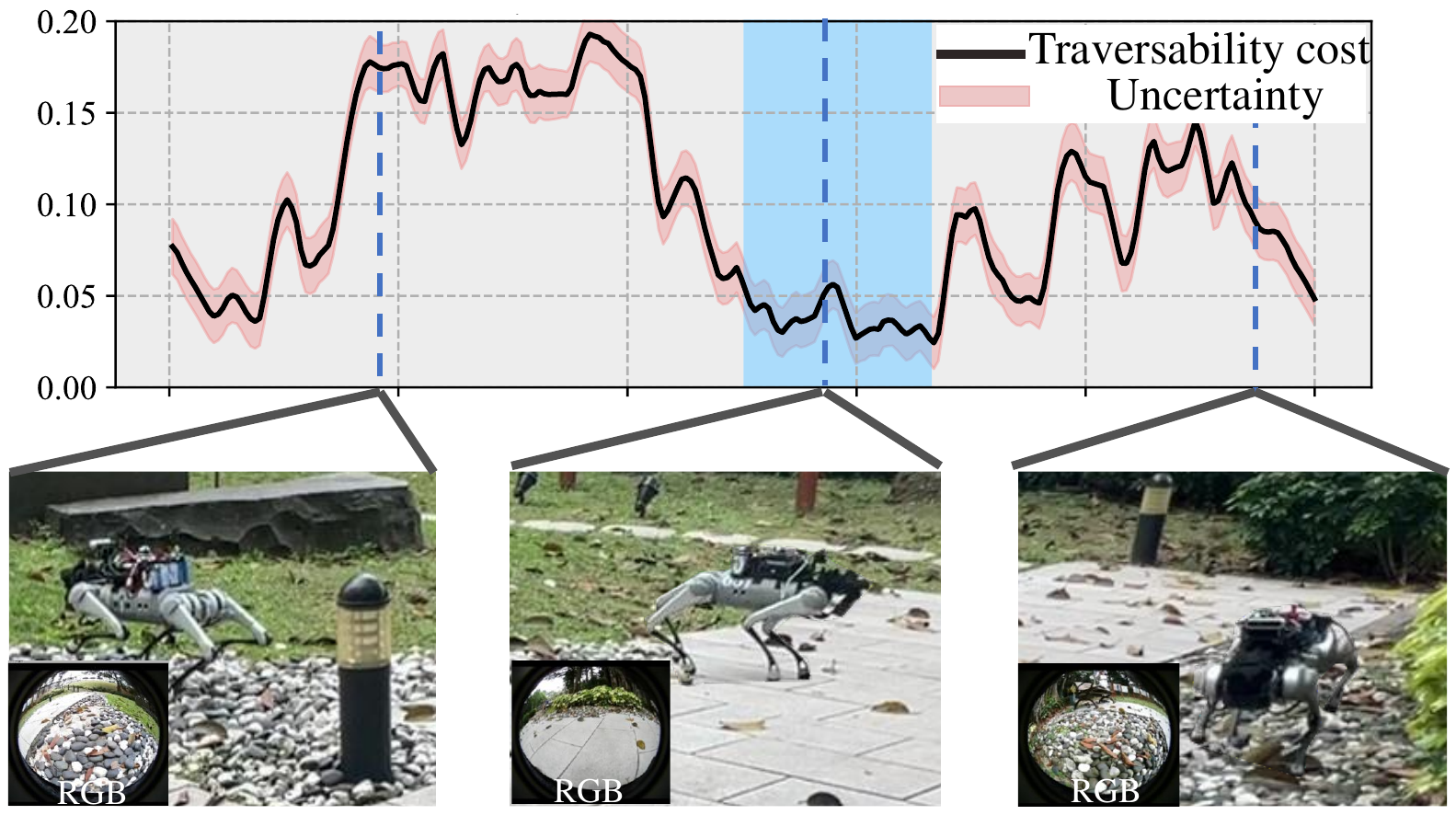}
\caption{Traversability cost estimation results (black curve) with uncertainty (red block) is shown in this figure when the robot is walking on a cross area of cobblestone roads and flat roads. During the time slots when the robot is walking on the cobblestone area, the estimated result has a high traversability cost, while on the flat roads, the estimated result shows a lower value close to zero (the blue area of the plot), which matches the real observation that the robot base shakes when slipperiness occurs between the foot and rocks when walking on the cobblestone area.}
\label{traver_single}
\vspace{-0.3cm}
\end{figure}

\begin{figure}[!t]
\centering
\includegraphics[width=3.4in]{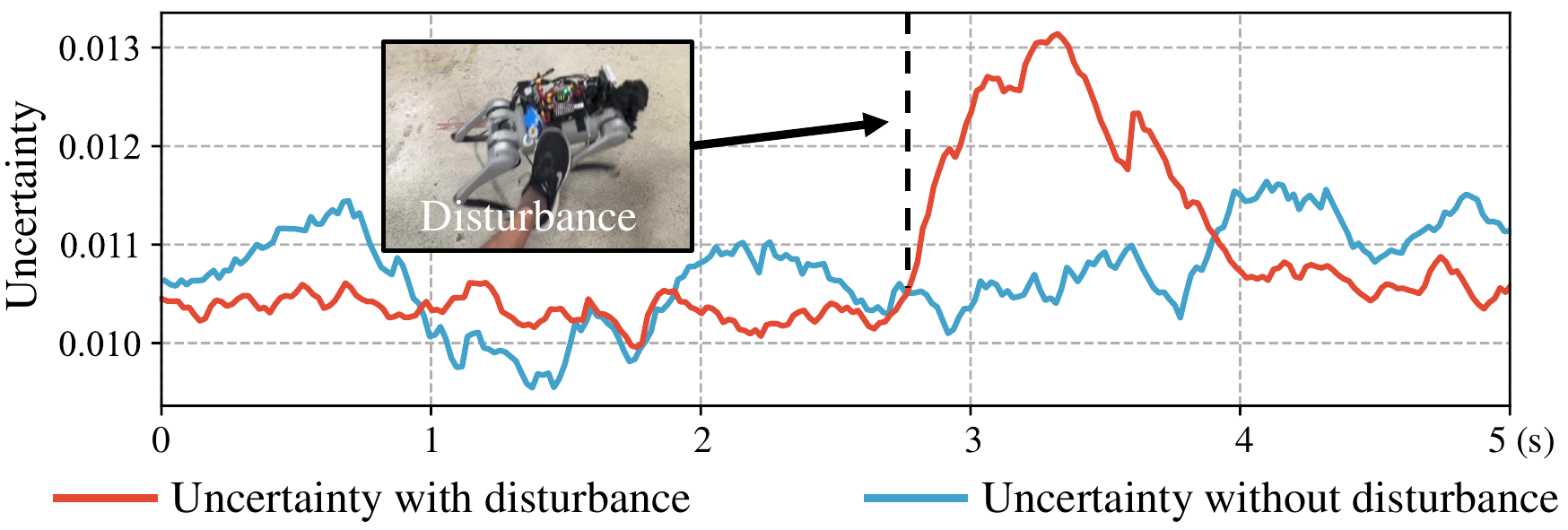}
\caption{The change of the uncertainty level when large disturbances that are not accurately simulated happen. The uncertainty level \(\sigma\) of the estimated traversability $T_{\text{value}}$ represented by the orange line rises significantly after the disturbance occurs and then returns to a nominal level once the robot recovers from the disturbance. For comparison, we also plot the blue line representing the nominal walking situation on cobblestone terrain without disturbances, where the overall uncertainty level is lower.}
\label{uncertainty_comparison}
\vspace{-0.3cm}
\end{figure}

\begin{figure*}[!t]
\centering
\includegraphics[width=0.99\linewidth]{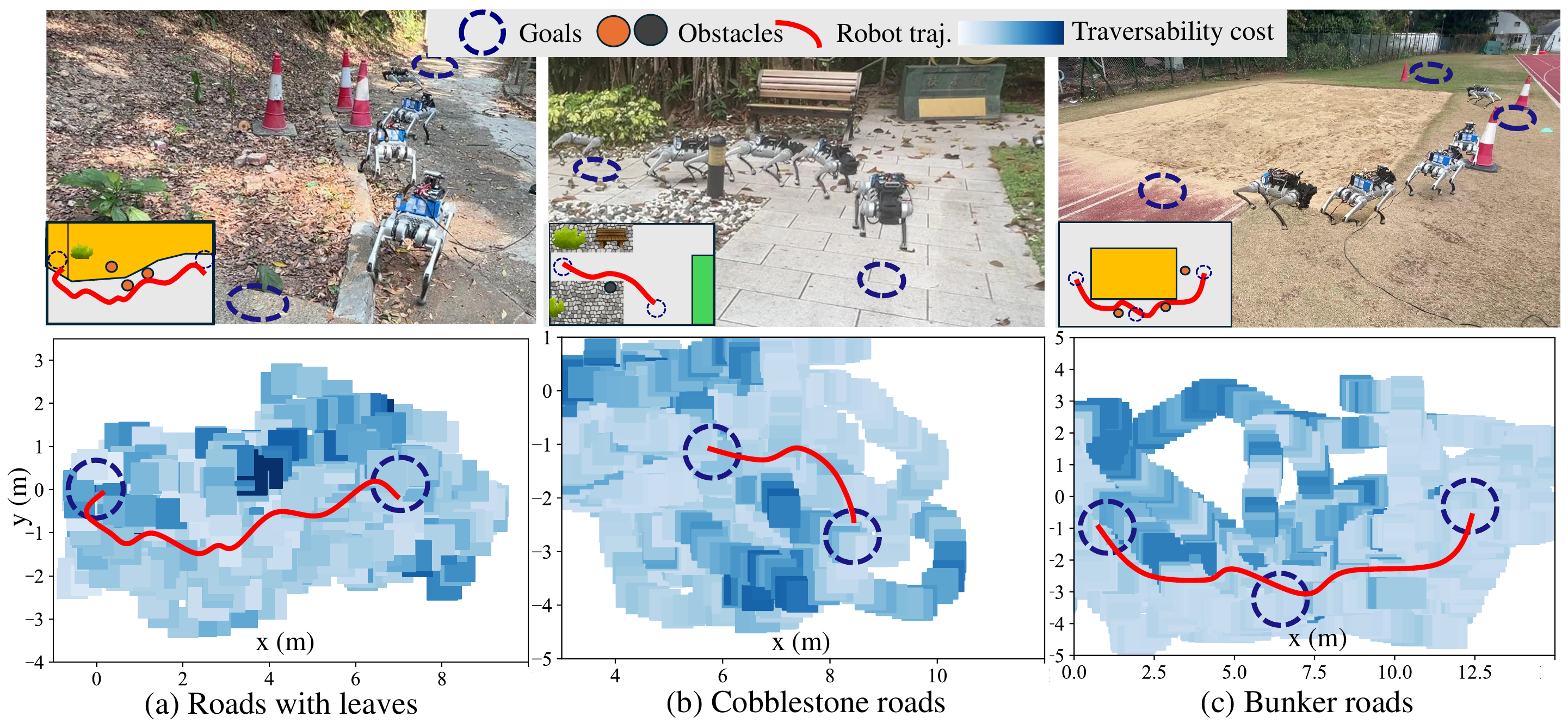}
\caption{Snapshots of the robot trajectories during the evaluation of the planner policy $\pi^p_{\text{rgbd}}$ are shown. The robot successfully learns to avoid obstacles, avoid areas with high traversability cost (areas with leaves, areas full of cobblestones, and areas of the sand-filled bunker), and reach goals (marked as dotted circles), with red curves representing the robot's trajectories. Simplified terrain schematics are provided to illustrate the top-down view of each environment setting. The second line of this figure showcases the traversability estimation results represented with deep (high traversability cost) and light(low traversability cost) blue shades using the data collected for the whole training process. The robot trajectories avoid most of the hard-to-traverse areas represented by darker blue shades as a desired behavior.}
\label{real_ps_robts}
\vspace{-0.4cm}
\end{figure*}

As shown in Fig.~\ref{traver_single}, the estimation results accurately reflect changes in terrain. The final result of the scalar traversability cost $T$ significantly increases when the robot traverses cobblestone areas and decreases on nominal concrete roads, correctly reflecting the difficulty of traversing through these regions. For example, cobblestone areas are challenging for the robot to traverse and introduce obvious oscillation on the robot base.
The results of this estimatio is used as part of the reward $r^p_t$ in~\eqref{rewards2} to train the planner \(\pi_{\text{rgbd}}^p\).

Furthermore, as shown in the second row of Fig.~\ref{real_ps_robts}, we visualize the estimated traversability cost collected during the training process by representing the level of traversability with varying shades of blue (darker blue shades indicate higher traversability cost). 
We overlap the traversability estimation sample points directly onto the ground truth position points from a top-down view. As shown in the figure, the areas with darker blue shades, representing high traversability cost, coincide with the actual difficult-to-traverse areas (such as piles of leaves, cobblestone regions, and the sand-filled bunker), demonstrating the reliability and the accuracy of the proposed traversability estimator. It's important to note that we do not rely on explicit mapping from traversability to position or a prebuilt map labeled by estimated traversability during training.

\subsubsection{Uncertainty-aware Estimation}
To emphasize the necessity of explicit \textit{uncertainty} evaluation for the estimation $T_{\text{value}}$ described in Sec.~\ref{subsec:estimate}, we examine the uncertainty level of the traversability estimator under scenarios that are not accurately simulated, \textit{i.e.} a large disturbance. When this large disturbance happens, the estimation from $V^c_{\text{terrain}}$ becomes uncertain about itself. As shown in Fig.~\ref{uncertainty_comparison}, the uncertainty level \(\sigma\) represented by the orange line rises significantly after the disturbance occurs and then returns to a nominal level once the robot recovers from the disturbance. We also plot the level of uncertainty during nominal walking without disturbance when the uncertainty level is acceptably low and the estimation $T_{\text{value}}$ is reliable, represented by blue lines. To have a reliable estimation, when the uncertainty level of $T_{\text{value}}$ rises, less confidence is placed on $T_{\text{value}}$ and the confidence on the backup estimation $T_{\text{track}}$ from velocity tracking errors increases as described in~\eqref{trav_eq}.

\subsection{Traversability-Aware Navigation}

We then evaluate the trained planning policy $\pi^p_{\text{rgbd}}$ using the proposed traversability estimator. 
The robot's actual trajectories, after training coverages, are recorded as red lines in Fig.~\ref{real_ps_robts}.

\begin{itemize}
    \item \textbf{Roads with Leaves} As shown in Fig.~\ref{real_ps_robts}(a), the robot learns to avoid the leaf-covered region, which is challenging to traverse due to the piles of leaves and branches that can cause the robot's legs to get stuck and lead to instability. 
    Instead of choosing the fastest route from the robot's starting point to the goal \textit{i.e.}, a straight line across the leaf-covered region, the quadruped learns to identify the leaf-covered road as a high traversability cost area from visual input and avoids the region earlier before entering it. This is done by training with our proposed traversability-aware navigation framework.
    Additionally, the robot successfully learns to avoid obstacles and reach goals as specified by the general robot navigation tasks.
    \item \textbf{Cobblestone Roads} In the second scenario depicted in Fig.~\ref{real_ps_robts}(b), the difficult-to-traverse area consists of cobblestones, which create slippery conditions and instability for the quadruped. Using our proposed method, the robot learns to identify this low traversability area from color input and chooses to navigate around the cobblestone areas to reach the goal, rather than directly crossing over them which could lead to a shorter but high traversability cost path.
    \item \textbf{Bunker Roads} In Fig.~\ref{real_ps_robts}(c), we see a clear boundary between the bunker sand area and the flat area on the plot of the estimated high/low traversability cost. 
    The robot learns to avoid such a sand-filled bunker. 
    However, the robot does not adopt a conservative strategy where it leaves the boundary of the bunker too far. Instead, it walks just near the boundary. This is due to our training strategy, which integrates the robot's own experiences to learn from a combination of reward signals.
    \end{itemize}
In summary, the results demonstrate that the robot-centric traversability estimation from the locomotion system provides reliable signals for the planner to generate trajectories with lower traversability costs. The success of the real-world experiment validates that the robot learns to bind the color information with the level of traversability estimated from its own experiences, enabling it to avoid difficult-to-traverse regions based on visual input.

\subsection{Generalization to Unseen Scenarios}
To further validate the generalization of the trained policy $\pi_{\text{rgbd}}^p$ in the real world, we evaluate the policy over two kinds of scenarios where the robot is not trained: (1) dynamic obstacle avoidance and (2) unseen surroundings.

\subsubsection{Dynamic Obstacles}
Instead of using static obstacles as used during training, we introduce walking persons as dynamic obstacles, which is commonly encountered in areas with pedestrian traffic.
After training and converging the policy in each scenario, we test its ability to handle dynamic obstacles by introducing a person randomly walking in the robot's path. 
The results are recorded in the supplementary video.

\subsubsection{Unseen Environment Surroundings} 
\begin{figure}[!t]
\centering
\includegraphics[width=.99\linewidth]{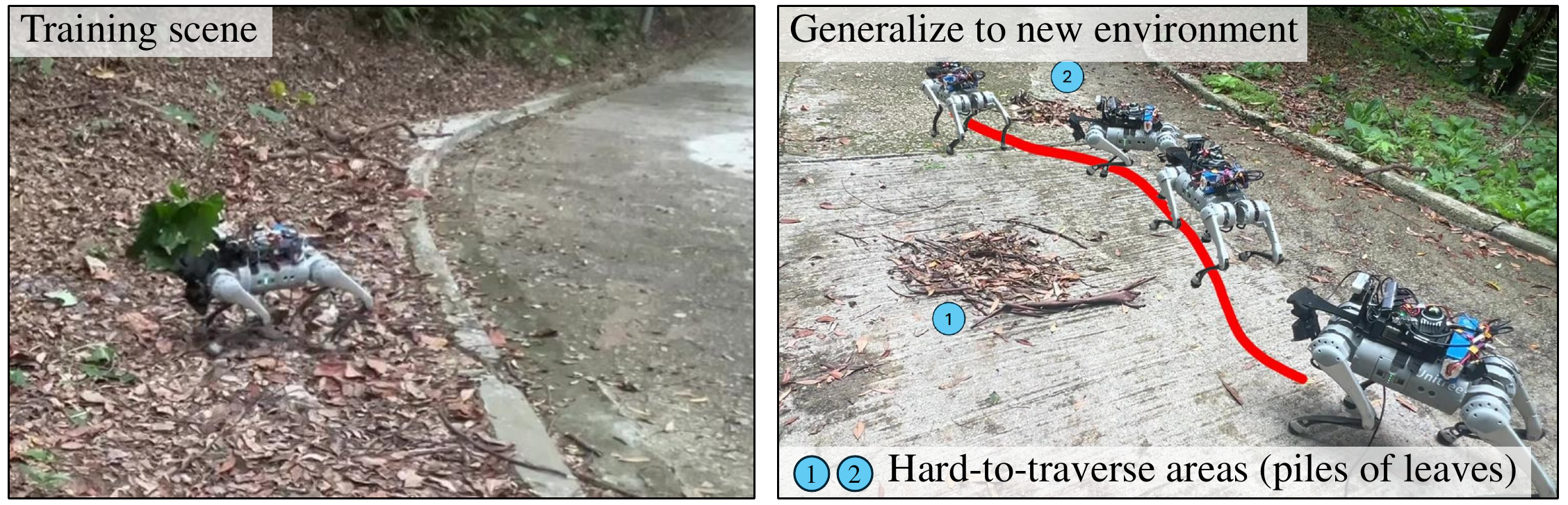}
\caption{The trained policy \(\pi_{\text{rgbd}}^p\) on Fig.~\ref{real_ps_robts}(a) generalize successfully to the new environment over unseen terrains which is different from the training scenes. Two areas that are full of leaves marked with serial numbers 1 and 2 are set to represent the hard-to-traverse areas, which are unseen layouts for the planner \(\pi_{\text{rgbd}}^p\).}
\label{general-blocks}
\vspace{-0.3cm}
\end{figure}

We also assess the generalization ability of the trained policy over the environment surroundings that differ from the one during training. 
In this test, we took the trained policy in experiment \textit{Roads with leaves} (Fig. 9(a)) into a new environment as shown in Fig.~\ref{general-blocks} to testify the generalization ability. Unlike the training environment, where leaves are cluttered over a large area on one side of the concrete road, we randomly place small piles of leaves on the road. As demonstrated in Fig.~\ref{general-blocks}, without any further fine-tuning, the robot is able to avoid the placed piles of leaves and reach the goal. 
This experiment highlights that the trained planner is not merely overfitting to a specific environment but has learned to generalize unseen environment surroundings, indicating its potential for application in more complex and large-scale outdoor scenarios.

\section{Conclusion and Future Work}
In conclusion, we have introduced a multi-stage hierarchical reinforcement learning framework to achieve traversability-aware RGBD-based quadrupedal navigation. The framework is designed to train the navigation planner using real-world multi-modal data and the robot's own experiences interacting with various terrains. 

The key design choices within the proposed hierarchical framework are thoroughly validated through extensive benchmarks.

Our approach offers a novel way to train vision-based navigation planners that consider robot-centric traversability, making it easy to apply and customize planners for different robot platforms in complex outdoor environments. However, one limitation of our approach is that it struggles to generalize to entirely new environments with different terrain textures. 

For instance, a planner trained on \textit{Roads with Leaves} might struggle with new terrain types, like a sand-filled bunker in \textit{Bunker Roads}.
To address this, a possible future direction could involve combining data collected from all different terrains into a unified offline dataset. 

Thanks to the flexible online traversability estimator, we can easily expand the offline dataset by collecting transition pairs as the robot navigates diverse terrains during routine operations. This extensive offline dataset can be used for zero-shot transfer with offline learning methods or adapted quickly to new scenarios using our proposed online learning approach with minimal rollouts.

\bibliographystyle{IEEEtranN}
{ \bibliography{bib/ref}}

\vspace{-1.2cm}
{
\footnotesize

\end{document}